\documentclass{./styles/svproc}
\usepackage{url}

\usepackage[numbers]{natbib}
\usepackage{enumitem}
\usepackage{amsmath}
\usepackage{newtxtext}
\usepackage{graphicx} \usepackage[font=small,labelfont=bf]{caption}
\usepackage[caption=false]{subfig}
\usepackage[varvw]{newtxmath}
\usepackage[bottom]{footmisc}
\usepackage[dvipsnames]{xcolor}
\DeclareMathOperator{\Tr}{Tr}
\newcommand{\ie}{i.e.,\ }
\newcommand{\eg}{e.g.,\ }

\begin{document}

\mainmatter              
\title{Decentralized Risk-Aware Tracking of Multiple Targets}
\titlerunning{Decentralized Risk-Aware Tracking of Multiple Targets}  
%
\author{Jiazhen Liu\inst{1} \and Lifeng Zhou\inst{1}
\and Ragesh Ramachandran\inst{2} \and Gaurav S. Sukhatme\inst{2}\thanks{G.S. Sukhatme holds concurrent appointments as a Professor at USC and as an Amazon Scholar. This paper describes work performed at USC and is not associated with Amazon.} \and Vijay Kumar\inst{1}}
\authorrunning{Jiazhen Liu et al.} 
%
%
\institute{University of Pennsylvania, Philadelphia PA 19104, USA \\
\email{jzliu, lfzhou, kumar@seas.upenn.edu}
\and
University of Southern California, Los Angeles CA 90007, USA \\
\email{rageshku, gaurav@usc.edu}}

\maketitle              

\begin{abstract}
We consider the setting where a team of robots is tasked with tracking multiple targets with the following property: approaching the targets enables more accurate target position estimation, but also increases the risk of sensor failures. Therefore, it is essential to address the trade-off between tracking quality maximization and risk minimization. In the previous work \cite{mayya2022adaptive}, a centralized controller is developed to plan motions for all the robots -- however, this is not a scalable approach. Here, we present a decentralized and risk-aware multi-target tracking framework, in which each robot plans its motion trading off tracking accuracy maximization and aversion to risk, while only relying on its own information and information exchanged with its neighbors. We use the control barrier function to guarantee network connectivity throughout the tracking process. Extensive numerical experiments demonstrate that our system can achieve similar tracking accuracy and risk-awareness to its centralized counterpart.
\end{abstract}
\section{Introduction}
Tracking multiple targets using a decentralized robot team finds applications in many settings, \eg tracking biochemical or nuclear hazards~\cite{schwager2017multi} or hot-spots in a fire~\cite{pham2017distributed}. In these settings, approaching the targets closely generally produces more accurate state estimates, but this comes at a cost -- the closer the team of robots comes to the targets, the more likely that they experience sensor failure or other system faults due to \eg radiation or heat emitted by the targets. In certain cases, the targets may even be actively hostile. The challenge is to maximize tracking accuracy while avoiding risky behavior on the part of the robot team. 
A common strategy for achieving team coordination is to devise a central controller to make team-level decisions for all robots~\cite{robin2016multi}. This relies on the assumption that all robots have a sufficiently large communication range to stay connected with a central station. In real applications, robots may have a limited communication range. A decentralized approach -- where robots only communicate with nearby team members -- is a natural choice~\cite{sung2020distributed, zhao2019systemic}. In a decentralized strategy, each robot makes its own decisions according to local information gathered by itself and its peers within a certain communication range. If such a decentralized strategy is to achieve the same level of task execution as a centralized system, it is essential for it to maintain network connectivity, such that robots could share information mutually and make the most use of their resources. 

Motivated by these issues, we present a decentralized risk-aware multi-target tracking framework. Our contributions are as follows: 
\begin{enumerate}
    \item \textbf{Our system is risk aware} -- it automatically trades off tracking quality and safety;
    \item \textbf{Our system is fully decentralized} -- each robot makes decisions relying on information sharing between neighbors within a communication range. We apply the control barrier function (CBF) in a decentralized manner to guarantee network connectivity maintenance and collision avoidance;
    \item \textbf{Our system produces comparable results to its centralized counterpart} -- we extensively evaluate the performance of our system through simulated experiments in Gazebo environment under various settings, and compare it to its centralized counterpart from our previous work~\cite{mayya2022adaptive}. We present qualitative demonstrations to provide an intuitive illustration. Quantitative results on metrics including tracking error and risk level are also exhibited.
\end{enumerate}

\section{Related Work}
Multi-robot multi-target tracking is usually formulated as an iterative state estimation problem~\cite{bar2004estimation,zhou2011multirobot,dames2017detecting,zhou2018active,zhou2019sensor,dames2020distributed}. Various methods have been proposed for this; most of them design different flavors of filters to reduce the uncertainty in target pose estimation during the tracking process~\cite{mayya2022adaptive,bar2004estimation,dias2013cooperative,wakulicz2021active}. In the more challenging scenario where the targets may adversarially cause damage to the robots, safety becomes a major concern when the robot team plans its motion. To incorporate risk-awareness into a multi-robot system,~\cite{wang2017safety} uses safety barrier certificates to ensure safety. With a special emphasis on tracking and covering wildfire, ~\cite{pham2017distributed} encodes hazards from fire using an artificial potential field. Mutual information is used in~\cite{schwager2017multi} to build a probabilistic model of the distribution of hazards in the environment.  

Considering that a multi-robot system controlled by a central planner does not scale well~\cite{mayya2022adaptive}, efforts have been made to decentralize the multi-target tracking task. For example,~\cite{sung2018distributed} adopts the local algorithm for solving the max-min problem to solve multi-target tracking in a distributed manner. In~\cite{tallamraju2018decentralized}, a decentralized MPC-based strategy is proposed that generates collision-free trajectories. In~\cite{ji2007distributed}, the general problem of multi-robot coordination while preserving connectivity is addressed. As it points out, a frequently adopted method to coordinate multiple robots in order to achieve a common goal is to use a consensus-based estimation algorithm.

To enable multi-robot multi-target tracking in a decentralized framework, it is typically required that the connectivity of the underlying network graph is maintained. To do this,~\cite{capelli2021decentralized} formulates the connected region as a safety set and uses a CBF to render it forward invariant such that the network will always remain connected as long as the robots are initialized properly. For a systematic elaboration of the theory of CBF and its relationship with Lyapunov control, see~\cite{ames2019control}. In~\cite{sabattini2013distributed}, a discussion of using graph spectral properties such as the algebraic connectivity to ensure that the network is connected is given. In~\cite{ji2007distributed}, the connectivity maintenance problem is approached by assigning weights to the edges and designing control strategies accordingly.

\section{Problem Formulation}
    \subsection{Notation}
    \label{sec:notation}
    We use capital letters in boldface (\eg $\mathbf{A}$) to denote matrices, lowercase letters in boldface (\eg $\mathbf{x}$) to denote vectors, and lowercase letters in regular font (\eg $q$) to denote scalars. $\mathbf{1}_{M \times N}$ denotes an all-one matrix of shape $M \times N$. 
    $\mathbf{I}_M$ is the identity matrix of shape $M \times M$. $\mathbf{M}_1 \otimes \mathbf{M}_2$ gives the Kronecker product of two matrices $\mathbf{M}_1$ and $\mathbf{M}_2$. $[\mathbf{v}_1; \mathbf{v}_2; \cdots; \mathbf{v}_n]$ represents vertical vector concatenation. The vertical concatenation of matrices or scalars are defined in the same fashion. $\text{Diag}([\mathbf{M}_1, \mathbf{M}_2, \cdots, \mathbf{M}_n])$ represents diagonal matrix construction with constituents $\mathbf{M}_1, \mathbf{M}_2, \cdots, \mathbf{M}_n$ placed along the diagonal. $\Tr(\mathbf{M})$ is the trace of matrix $\textbf{M}$. We use $\|\cdot \|$ to denote the 2-norm for vectors. For a set $\mathcal{S}$, $|\mathcal{S}|$ returns its cardinality. 
    $[z]$ denotes the set $\{1,2,\cdots, z\}$ for an arbitrary integer $z \in \mathbb{Z}^{+}$. $\mathscr{N}(\mathbf{0}, \mathbf{Q})$ represents the Gaussian distribution with a zero mean and covariance matrix $\mathbf{Q}$. 
    
    \subsection{Robot and Target Modeling}
    \label{sec:team_model}
    The multi-target tracking task is assigned to a team of $N$ robots, indexed by $\mathcal{R}:=[N]$. For each robot $i \in [N]$ with state $\mathbf{x}_{i}\in \mathcal{X} \subseteq \mathbb{R}^{p}$ and control input $\mathbf{u}_{i} \in \mathcal{U} \subseteq \mathbb{R}^{q}$, its motion is modeled by the control-affine dynamics: 
    \begin{equation}\label{eqn:robot_dyn}
       \dot{\mathbf{x}}_{i} = f(\mathbf{x}_{i}) + g(\mathbf{x}_{i})\mathbf{u}_{i}
    \end{equation}
    where $f:\mathbb{R}^p \rightarrow \mathbb{R}^p$ and $g: \mathbb{R}^p \rightarrow \mathbb{R}^{p\times q}$ are continuously differentiable vector fields. 
    The robot team is equipped with a set of sensors that are heterogeneous in terms of their sensing ability. Suppose there are in total $U$ different sensors indexed by $[U]$. We encode the measurement characteristic of sensor $k\in [U]$ using a vector $\mathbf{h}_k$. At time step $t$, we denote the sensor status, \ie whether each sensor is functioning or damaged, for robot $i$ as:
    \begin{equation}
        \label{eq:sensor_status}
        \mathbf{\Gamma}_{ki, t} = \begin{cases}
        1, \textnormal{if sensor $k$ on robot $i$ is functioning well at time $t$}; \\
        0, \textnormal{otherwise}. \\
        \end{cases}
    \end{equation}
    Based on this, $\mathbf{\Gamma}_t$ denotes the sensor status of the whole robot team at time $t$. The set of sensors available at time $t$ for robot $i$ is denoted as $\gamma_{i,t} = \{k\in [U] | \mathbf{\Gamma}_{ki, t} = 1\}$. Combining the sensor status with the measurement characteristic of each sensor, the measurement matrix for robot $i$ taking measurements of the target $j$ is: 
        $\mathbf{H}_{ij, t} = [\mathbf{h}_{\gamma_{i,t}(1)}; \mathbf{h}_{\gamma_{i,t}(2)}; \cdots;\mathbf{h}_{\gamma_{i,t}(|\gamma_{i,t}|)}]$, 
     and the measurement matrix for robot $i$ measuring all $M$ targets is: $\mathbf{H}_{i,t} = \mathbf{I}_M \otimes \mathbf{H}_{ij, t}$. 
    
    Without loss of generality, we assume that all the robots have a homogeneous maximum communication range $R_\text{comm}$, as heterogeneity in the robots' communication ability is beyond the scope of this paper and is left for future exploration. For robot $i$, we define the region where other robots could have information exchange with it as $\chi_i := \{\mathbf{x}\in \mathcal{X} | \| \mathbf{x} - \mathbf{x}_i \| \leq R_\text{comm}\}$. We call the robots that are within $\chi_i$ as robot $i$'s neighbors. A robot is only able to communicate with its neighbors. The list of indices of robot $i$'s neighbors is defined as $\mathcal{N}_i := \{j \in [N], j\neq i | \mathbf{x}_j \in \chi_i\}$. Note that $\mathcal{N}_i$ is time-varying for each robot $i\in \mathcal{R}$ as the robots actively reconfigure their positions to track the targets.
    
    Let the $M$ targets be indexed by $\mathcal{T} := [M]$. Denote the true state of target $j$ as $\textbf{z}_{j, t} \in \mathcal{X} \subseteq \mathbb{R}^p$, and its control input as $\textbf{v}^{}_{j, t} \in \mathcal{V} \subseteq \mathbb{R}^q$, both at time step $t$. The dynamics of the targets follow: 
    \begin{equation}
        \mathbf{z}_{t+1} = \mathbf{A} \mathbf{z}_{t} + \mathbf{B} \mathbf{v}_{t} + \mathbf{w}_{t}
    \end{equation}
    where $\mathbf{A}$ and $\mathbf{B}$ are the process and control matrices, respectively. $\mathbf{z}_t$ and $\mathbf{v}_t$ are obtained by stacking $\mathbf{z}_{j,t}$ and $\mathbf{v}_{j,t}, \forall j \in \mathcal{T}$ vertically. $\mathbf{w}_{t}\sim\mathscr{N}(\mathbf{0}, \mathbf{Q})$ is the process Gaussian noise with covariance matrix $\mathbf{Q}$. $\mathbf{A}, \mathbf{B}$ and $\mathbf{Q}$ are assumed to be known to the robots. Robot $i$'s measurement of the targets with noise $\boldsymbol{\zeta}_{i, t} = [\mathbf{\zeta}_{i1, t}; \mathbf{\zeta}_{i2, t}; \cdots; \mathbf{\zeta}_{iM, t}]$ is modeled by: 
    \begin{equation}
        \mathbf{y}_{i,t} = \mathbf{H}_{i, t} \mathbf{z}_{t} + \boldsymbol{\zeta}_{i,t}.
    \end{equation}
   We let $\mathbf{\zeta}_{ij,t} \sim \mathscr{N}(\mathbf{0}, \mathbf{R}_{ij, t}), \forall j\in \mathcal{T}$ where the measurement noise covariance matrix $\mathbf{R}_{ij, t}$ grows exponentially as the distance between robot $i$ and target $j$ increases,
   \begin{equation}
       \mathbf{R}^{-1}_{ij, t} = \text{Diag}([\mathbf{R}^{-1}_{ij1,t},\mathbf{R}^{-1}_{ij2,t},\cdots,\mathbf{R}^{-1}_{ij|\gamma_{i,t}|,t}])
   \end{equation}
   with $\mathbf{R}^{-1}_{ijk,t}$ relying on a particular sensor $k\in \gamma_{i, t}$ being:
   \begin{equation}
       \mathbf{R}^{-1}_{ijk, t} = w_k \text{exp}(-\lambda_k \|\mathbf{x}_{i,t}-\mathbf{z}_{j,t}\|)
   \end{equation}
   where $w_k$ and $\lambda_k$ are parameters characterizing the noise of sensor $k \in \gamma_{i,t}$. The target risk field $\phi$, which encodes the probability that a robot experiences sensor failures induced by the targets, inherits the same definition as our previous work~\cite[Sec. \textrm{III}-E, \textrm{V}]{mayya2022adaptive}. In particular, the risk from target $j$ acting on all robots is: 
   \begin{equation}
       \phi_j(\mathbf{x}) = \frac{c_j}{2\pi |\mathbf{\Sigma}_j|}\text{exp}(-\frac{1}{2}(\mathbf{x} - \mathbf{z}_j)^\texttt{T}\mathbf{\Sigma}_j (\mathbf{x} - \mathbf{z}_j))
   \end{equation}
    where $c_j$ and $\mathbf{\Sigma}_j$ denote the peak value and covariance for the risk field of target $j$, respectively. Correspondingly, the safety field is defined as $\pi_j(\mathbf{x}) = 1-\phi_j(\mathbf{x}), \forall j \in \mathcal{T}$.
    
    \subsection{Problem Definition}
    \label{sec:setup}
    We focus on planning the motion of each robot in the team to obtain an accurate estimation of the targets' positions while at the same time avoiding risk, in a decentralized manner at each time step $t$ throughout the tracking process. Specifically, for each robot $i \in \mathcal{R}$, the goal is to solve the following optimization program: 
    \begin{subequations}
        \label{general-problem}
        \begin{align}
        \min_{\mathbf{x}_i, \boldsymbol{\delta}_{i1}, \delta_{i2}} \quad & q_1 \eta_i \left\lVert \boldsymbol{\delta}_{i1}\right\rVert^2 + q_2 \frac{1}{1+\eta_i}\delta_{i2}^2 \label{eq:general-cost} \\
        \textrm{s.t.} \quad 
      & \Tr(\mathbf{P}_{ij}) \leq \boldsymbol{\rho}_{i1, j} + \boldsymbol{\delta}_{i1, j},\enspace \boldsymbol{\delta}_{i1, j}\geq 0,\enspace \forall j\in \mathcal{T}; \label{eq:general-perf} \\
      & \Tr(\mathbf{O}_{i, \mathbf{\Pi}}^{-1}) \leq \rho_{i2} + \delta_{i2},\enspace\delta_{i2} \geq 0;  \label{eq:general-risk} \\
      & |\mathcal{N}_i| > 0; \label{eq:general-connectivity} \\
      & \left\lVert \mathbf{x}_i - \bar{\mathbf{x}}_{i}\right\rVert \leq d_{\max} \label{eq:general-dynamics}; \\
      & \left\lVert \mathbf{x}_i - \mathbf{x}_l\right\rVert \geq d_{\min},\enspace \forall l \in \mathcal{R}, l \neq i. \label{eq:general-collision avoidance}
    \end{align}
    \end{subequations}
    The time index $t$ is dropped here for the brevity of notation. $\mathbf{x}_i$ is the optimization variable representing robot $i$'s position at the next step, while $\bar{\mathbf{x}}_i$ refers to robot $i$'s current position. The cost function Eq.~(\ref{eq:general-cost}) balances competing objectives of high tracking quality (first term) and safety (second term), using the sensor margin $\eta_i$ which encodes the abundance of sensors available for robot $i$. The larger the value of $\eta_i$, the more powerful the sensor suite on robot $i$ is for observing the targets, and therefore the first term in the objective is more stressed. On the contrary, when the sensors on robot $i$ are limited, $\eta_i$ becomes small, enforcing the second term in Eq.~\ref{eq:general-cost} to be stressed prioritizing safety. $q_1, q_2$ are constants weighing the two objectives. The tracking error is represented using the trace of target estimation covariance matrix $\mathbf{P}_{ij}$ and is upper bounded by threshold $\boldsymbol{\rho}_{i1,j},j\in\mathcal{T}$ to ensure tracking accuracy, as in Eq.~(\ref{eq:general-perf}). Meanwhile, the risk of sensor failure is encoded by $\Tr(\mathbf{O}^{-1}_{i, \mathbf{\Pi}})$, where $\mathbf{O}_{i, \mathbf{\Pi}}$ is the risk-aware observability Gramian for robot $i$, defined similarly as the ensemble form in~\cite[Eq. (14)]{mayya2022adaptive}. 
    The risk level is upper bounded by a pre-specified threshold $\rho_{i2}$ to discourage highly risky behavior, as in Eq.~(\ref{eq:general-risk}). $\boldsymbol{\delta}_{i1,j}, \forall j \in \mathcal{T}$ and $\delta_{i2}$ are slack variables, which enable the ``\textit{bend but not break}'' behavior when sensor failures render it impossible to meet the strict constraints of $\boldsymbol{\rho}_{i1,j}, \forall j \in \mathcal{T}$ and $\rho_{i2}$. See~\cite[Sec. \textrm{IV}-B, C, D]{mayya2022adaptive} for more details on the design of sensor margin and observability Gramian. Moreover, according to Eq.~(\ref{eq:general-connectivity}), each robot should have at least one neighbor, so that the communication network formed by the robots remains connected. Note that the connectivity relationship is determined using $\mathbf{x}_i$ rather than $\bar{\mathbf{x}}_i$, to ensure that the robots are still connected at the next time step. Constraint Eq.~(\ref{eq:general-dynamics}) specifies the maximum distance $d_{\max}$ that robot $i$ could travel between two consecutive steps. Finally, Eq.~(\ref{eq:general-collision avoidance}) is the collision avoidance constraint to guarantee the minimum safety distance $d_{\min}$ between two robots. Notice that the optimization program in Eq.~(\ref{general-problem}) is a non-convex problem that is computationally difficult to solve. We discuss an approach to solve it in two stages in the next section. 
    
\section{Approach}
\begin{figure}[ht]
    \centering
        \includegraphics[width=0.8\textwidth]{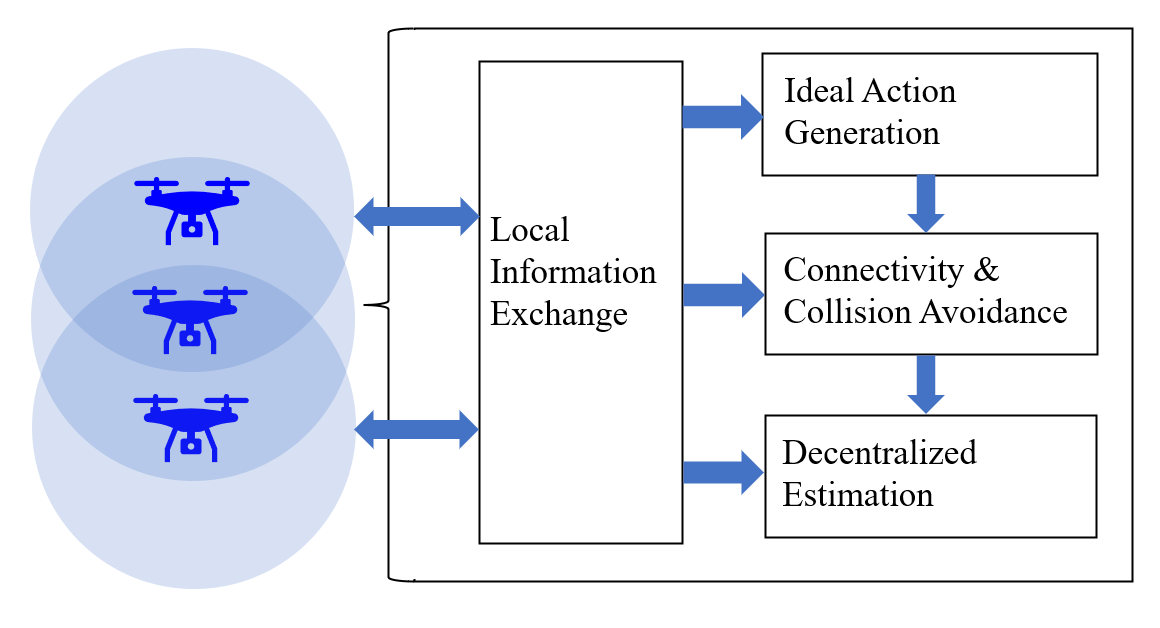}
        \caption{The proposed decentralized framework. Each robot performs the decentralized target tracking task using 4 modules: Local Information Exchange, Decentralized Ideal Action Generation, Decentralized Connectivity and Collision Avoidance, and Decentralized Estimation. The Local Information Exchange module takes charge of all communications between neighbors. }
        \label{fig:workflow}
    \end{figure}

    In the proposed decentralized system, every robot relies on four major modules to perform the target tracking task, as illustrated in Fig.~\ref{fig:workflow}. At each time step $t$, every robot first generates its ideal goal state that it wants to reach without considering constraints including network connectivity and collision avoidance. Its decisions are only based on trading off tracking quality and risk aversion. This is performed by the Ideal Action Generation module, as introduced in Sec.~\ref{sec:opt1}. Sec.~\ref{sec:qp} corresponds to the Connectivity and Collision Avoidance module, which employs CBF to ensure network connectivity and to avoid inter-robot collisions when the robots move towards their respective goal positions. After each robot moves to its new position, the Decentralized Estimation module generates a new estimate of target positions and maintains common knowledge of such estimate among neighbors using the Kalman-Consensus Filtering algorithm in~\cite{liu2017kalman}. We introduce it in Sec.~\ref{sec:ekf}. Note that the Local Information Exchange module is responsible for information sharing within the neighborhood and all the other three modules rely on it. 
    
    \subsection{Decentralized Goal Position Generation}
    \label{sec:opt1}
    In this section, we focus on computing the ``ideal'' positions that robots want to reach at every time step, only out of the desire for higher tracking accuracy and aversion to overly risky behavior, without considering the connectivity maintenance and collision avoidance constraints. To this end, each robot $i \in \mathcal{R}$ solves a non-convex optimization program as follows:  
    \begin{subequations}
        \label{non-convex-opt}
        \begin{align}
        \min_{\mathbf{x}_i, \boldsymbol{\delta}_{i1}, \delta_{i2}} \quad & q_1 \eta_i \left\lVert \boldsymbol{\delta}_{i1}\right\rVert^2 + q_2 \frac{1}{1+\eta_i}\delta_{i2}^2 \label{eq:cost} \\
        \textrm{s.t.} \quad 
      & \text{Tr}(\mathbf{P}_{ij}) \leq \boldsymbol{\rho}_{i1, j} + \boldsymbol{\delta}_{i1, j},\enspace \boldsymbol{\delta}_{i1, j}\geq 0,\enspace \forall j\in \mathcal{T}; \label{eq:perf}\\
      & \text{Tr}(\mathbf{O}_{i, \mathbf{\Pi}}^{-1}) \leq \rho_{i2} + \delta_{i2},\enspace \delta_{i2} \geq 0.  \label{eq:risk-aware}
    \end{align}
    \end{subequations}
    This is a less constrained version of the optimization program in Eq.~(\ref{general-problem}). Note that to achieve team-level coordination, the computation of $\mathbf{P}_{ij}$ and $\mathbf{O}_{i, \mathbf{\Pi}}^{-1}$ requires not only robot $i$'s information, but also that from its neighbors acquired through the Local Information Exchange module. Since each robot runs its own optimization program, all robots in the team are required to solve the optimization program in Eq.~(\ref{non-convex-opt}) over multiple rounds, until their ideal goal positions converge.  

    \subsection{Decentralized Motion Synthesis}
    \label{sec:qp}
    Given the generated ideal goal positions, we aim to steer the robots towards their goal positions while preserving the network connectivity and inter-robot safety distance. We achieve this by using the control barrier functions. 

    We consider the robots with control-affine dynamics $\dot{\mathbf{x}} = f(\mathbf{x}) + g(\mathbf{x}) \mathbf{u}$ as defined in Eq.~(\ref{eqn:robot_dyn}), where $f(\mathbf{x}), g(\mathbf{x})$ are locally Lipschitz functions. We let $\mathcal{C}$ be the region where we want the robot team to stay within. Concretely, we define $\mathcal{C}$ as the superlevel set of a continuously differentiable function $h(\mathcal{X})$: $\mathcal{C} = \{\mathcal{X}\in \mathbb{R}^p | h(\mathcal{X}) \geq 0 \}$. From~\cite{ames2019control}, we know that if we can select a feasible  $\mathbf{u} \in \mathbb{R}^q$ such that
    \begin{equation}
        L_f h(\mathcal{X}) + L_g h(\mathcal{X})\mathbf{u} + \alpha(h(\mathcal{X})) \geq 0
    \end{equation}
    is satisfied, we could render $\mathcal{C}$ forward invariant and ensure that if the robots initially start from an arbitrary state inside $\mathcal{C}$, they will always remain within $\mathcal{C}$. Note that $L_f$ and $L_g$ represent the Lie derivatives of $h(\cdot)$ along the vector fields $f$ and $g$, respectively.
    $\alpha(\cdot)$ is an extended class $\mathcal{K}$ function as introduced in~\cite{capelli2021decentralized}. 
    
    Back to our case, to ensure network connectivity, we define $h(\cdot)$ as:
    \begin{equation}
        h(\mathbf{x}) = \lambda_2 - \epsilon
    \end{equation}
    where $\lambda_2$ is the algebraic connectivity, which is the second smallest eigenvalue of the weighted Laplacian matrix of the robots' communication graph, and $\epsilon$ is the connectivity threshold adjusting the extent to which we want to enforce this connectivity constraint. A larger $\epsilon$ ensures network connectivity more strictly. The weighted Laplacian matrix is: $\mathbf{L}_{t} = \mathbf{D}_t - \mathbf{A}_t$. $\mathbf{D}_t$ and $\mathbf{A}_t$ represent the weighted degree matrix and weighted adjacency matrix, respectively. Specifically, we adopt the same definition from~\cite[Eq. (5)]{capelli2021decentralized} to define the adjacency $a_{il, t}$ between robot $i$ and $l$ at time $t$, using the distance $d_{il,t}$ between them. That is, 
    \begin{equation}
        \label{eq:adjacency}
        a_{il,t} = \begin{cases}
          e^{(R^2_\text{comm} - d_{il,t}^2)^2 / \sigma}-1, & \text{if $d_{il,t} \leq R_\text{comm}$};\\
          0, & \text{otherwise}, 
        \end{cases}
    \end{equation}
    where $\sigma \in \mathbb{R}$ is a normalization constant. Based on these, we formulate the connectivity maintenance constraint using CBF in Eq.~(\ref{eq:qp_connect}) and similarly formulate the collision avoidance constraint in Eq.~(\ref{eq:qp_collision_avoid}). With these constraints, each robot $i\in\mathcal{R}$ is steered towards its desired position by solving the following quadratically constrained quadratic program:
    \begin{subequations}
        \label{eq:qp}
        \begin{align}
        \min_{\mathbf{u}_i} \quad & \frac{1}{2} \left\lVert \mathbf{u}_i - \mathbf{u}_{\text{des},i} \right\rVert^2 \label{eq:qp_cost}\\ 
        \textrm{s.t.} \quad & \left\lVert \mathbf{u}_i \right\rVert \Delta t \leq d_{\max}; \label{eq:qp_dynamics}\\
        & \frac{\partial \lambda_2}{\partial\bar{\mathbf{x}}_i}\mathbf{u}_i + \lambda_2 - \epsilon \geq 0; \label{eq:qp_connect}\\
        & \frac{\partial d^2_{il}}{\partial\bar{\mathbf{x}}_i}\mathbf{u}_i + d^2_{il} - d^2_{\min} \geq 0,\enspace \forall l \in \mathcal{N}_i, \label{eq:qp_collision_avoid}
        \end{align}
    \end{subequations}
    where $\mathbf{u}_{\text{des},i}$ denotes the ideal control input which could be computed easily using the desired ideal goal position generated in Sec.~\ref{sec:opt1}. $\Delta t$ is the time interval between two consecutive steps. $\bar{\mathbf{x}}_i$ is robot $i$'s state at the current step. $d_{il}$ is the distance between robot $i$ and $l$ calculated using their current states $\bar{\mathbf{x}}_i$ and $\bar{\mathbf{x}}_l$. The cost function in Eq.~(\ref{eq:qp_cost}) minimizes the discrepancy between robot $i$'s actual control input and its ideal one. Constraint Eq.~(\ref{eq:qp_dynamics}) specifies the maximum distance $d_{\max}$ that each robot could move between consecutive steps. 
    Moreover, Eqs.~(\ref{eq:qp_connect}) and~(\ref{eq:qp_collision_avoid}) are the connectivity maintenance and collision avoidance constraints, respectively, as introduced above.  
    
    In the decentralized system, every robot solves the optimization program in Eq.~(\ref{eq:qp}) to generate its actual control input. Similar to Sec.~\ref{sec:opt1}, we adopt an iterative procedure where all the robots exchange information and solve the optimization program in Eq.~(\ref{eq:qp}) repeatedly, until their solutions $\mathbf{u}_i, \forall i\in \mathcal{R}$ converge. 
 
    From the computational perspective, each robot $i\in \mathcal{R}$ needs to know $\lambda_2$ and $\frac{\partial \lambda_2}{\partial \bar{\mathbf{x}}_i}$ for calculating Eq.~(\ref{eq:qp_connect}). In a centralized system, a central controller has access to information of every robot and can directly compute $\lambda_2$ and $\frac{\partial \lambda_2}{\partial \bar{\mathbf{x}}_i}$. However, in the decentralized system, the robots need to compute them using local information. Inspired by~\cite[Sec. \textrm{III}-\textrm{IV}]{yang2010decentralized}, we utilize the Decentralized Power Iteration (PI) algorithm, which enables each robot to compute a local approximation of $\lambda_2$ and $\frac{\partial \lambda_2}{\partial \bar{\mathbf{x}}_i}$. Such approximation is achieved through repeated local information exchange. Specifically, in each round of the information exchange, the message passed between neighbors is: 
    \begin{equation}
        \label{eq:info_exchange}
        \mathcal{E}_i = \{ \nu_i, y_{i, 1}, w_{i, 1}, y_{i, 2}, w_{i, 2}\}
    \end{equation}
    where $\nu_i$ approximates the $i^{th}$ element of the eigenvector $\mathbf{\nu}$ corresponding to $\lambda_2$ of the weighted Laplacian matrix $\mathbf{L}$. $y_{i,1}$ is robot $i$'s estimate of $\text{Ave}(\{ \mathcal{\nu}_i\}),\forall i \in \mathcal{R}$, \ie average of all entries in $\mathcal{\nu}$. Similarly, $y_{i,2}$ is robot $i$'s estimate of $\text{Ave}(\{ \mathcal{\nu}_i^2\}),\forall i \in \mathcal{R}$, \ie the average of squares of all entries in $\mathcal{\nu}$. $w_{i,1}$ and $w_{i,2}$ are variables introduced for computation purpose. 
    
    \cite{yang2010decentralized} shows that the decentralized PI algorithm converges in a few iterations. Once it converges, every robot computes its local estimate of the algebraic connectivity $\lambda_2$ as: 
    \begin{equation}
        \lambda_2^{i} = - \frac{\sum_{l\in \mathcal{N}_i}L_{il}\nu_l}{\nu_i}
    \end{equation}
    Additionally, from~\cite[Eq. (8)]{capelli2021decentralized},  $\frac{\partial \lambda_2}{\partial \bar{\mathbf{x}}_i}$ is computed as:
    \begin{equation}
        \frac{\partial \lambda_2}{\partial \bar{\mathbf{x}}_i} = \sum_{l \in \mathcal{N}_i}\frac{\partial a_{il}}{\partial \bar{\mathbf{x}}_i}(\nu_i - \nu_l)^2
    \end{equation}
    where $a_{il}$ is the weighted adjacency introduced in Eq.~(\ref{eq:adjacency}). In this way, $\frac{\partial \lambda_2}{\partial \bar{\mathbf{x}}_i}$ is computed locally using the values of $\nu_i, \nu_l$ obtained from the decentralized PI algorithm. Substituting the approximations of $\lambda_2$ and $\frac{\partial \lambda_2}{\partial \bar{\mathbf{x}}_i}$ into optimization program in Eq.~(\ref{eq:qp}), each robot $i\in \mathcal{R}$ can solve Eq.~(\ref{eq:qp}) to obtain it actual control input.   
    
    \subsection{Decentralized Target Position Estimation}
    \label{sec:ekf}
    Whenever the robot team reaches a new state, each robot obtains new measurements of the targets and runs a Kalman Filter (KF) independently to generate new estimates of all targets' positions.
    Specifically, each robot $i \in \mathcal{R}$ computes the a priori estimate $\hat{\mathbf{z}}_{i,t|t-1}$ and covariance matrix $\mathbf{P}_{i,t|t-1}$ during the prediction step as:
    \begin{align}
        & \hat{\mathbf{z}}_{i,t|t-1} = \mathbf{A} \hat{\mathbf{z}}_{i,t-1|t-1},~
        & \mathbf{P}_{i, t|t-1} = \mathbf{A}\mathbf{P}_{i, t-1|t-1} \mathbf{A}^{\top} + \mathbf{Q},
    \end{align}
    where $\hat{\mathbf{z}}_{i, t-1|t-1}$ and $\mathbf{P}_{i, t-1|t-1}$ represent robot $i$'s estimate of the targets' positions and the corresponding covariance matrix at the previous time step, respectively. $\mathbf{A}$ is the process matrix for targets, and $\mathbf{Q}$ is the targets' process noise covariance matrix, both introduced in Sec.~\ref{sec:team_model}. 
    
    After robot $i\in \mathcal{R}$ obtains a new measurement $\mathbf{y}_{i,t}$ of the targets, the measurement residual $\Tilde{\mathbf{y}}_{i,t}$ and Kalman gain $\mathbf{K}_{i,t}$ are calculated as:
    \begin{align}
        &\Tilde{\mathbf{y}}_{i,t} = \mathbf{y}_{i,t} - \mathbf{H}_{i,t}\hat{\mathbf{z}}_{i, t|t-1},~
        &\mathbf{K}_{i,t} = \mathbf{P}_{i,t|t-1}\mathbf{H}_{i,t}^{\top}\mathbf{S}_{i,t}^{-1},
    \end{align} 
    where $\mathbf{S}_{i,t} = \mathbf{H}_{i,t}\mathbf{P}_{i, t|t-1}\mathbf{H}^{\top}_{i,t} + \mathbf{R}_{i,t}$, with $\mathbf{R}_{i,t}$ being the measurement noise covariance matrix as introduced in Sec.~\ref{sec:team_model}. The new measurement is incorporated into robot $i$'s estimation during the update step of KF:
    \begin{align}
        &\hat{\mathbf{z}}_{i, t} = \hat{\mathbf{z}}_{i, t|t-1} + \mathbf{K}_{i,t}\Tilde{\mathbf{y}}_{i,t},~
        &\mathbf{P}_{i,t} = \mathbf{P}_{i, t|t-1} - \mathbf{K}_{i,t} \mathbf{S}_{i,t}\mathbf{K}^{\top}_{i,t},
    \end{align}
    
    Since each robot runs its own KF, their estimates may be different. To maintain a common estimate of the targets' positions, we utilize the Kalman-Consensus Filtering algorithm~\cite[Table \textrm{I}]{liu2017kalman}. This is an iterative approach for the robots to reach an agreement on their estimates of the targets' positions. Each robot first exchanges the information matrix $\mathbf{\Omega}_{t}^{i}(0) = (\mathbf{P}_{i,t})^{-1}$ and information vector $\mathbf{q}_{t}^{i}(0) = (\mathbf{P}_{i,t})^{-1}\hat{\mathbf{z}}_{i,t}$ with its neighbors. Then, each robot $i \in \mathcal{R}$ performs multiple rounds of the following convex combination:
     \begin{align}
        &\mathbf{\Omega}_{t}^{i}(\tau + 1) = \sum_{l \in \mathcal{N}_i}\Tilde{\pi}_{i,l}\mathbf{\Omega}_{t}^{i}(\tau),~
        &\mathbf{q}_{t}^{i}(\tau + 1) = \sum_{l \in \mathcal{N}_i}\Tilde{\pi}_{i,l}\mathbf{q}_{t}^{i}(\tau),
    \end{align}
    until the values of $\mathbf{\Omega}_{t}^{i}$ and $\mathbf{q}_{t}^{i}$ converge. Note that $\Tilde{\pi}_{i,l}$ is the weight between robot $i$ and $l$, which is assigned using the method in~\cite{boyd2004fastest}. Then, each robot uses $\mathbf{\Omega}_{t}^{i}$ and $\mathbf{q}_{t}^{i}$ from the last round of the consensus algorithm to compute the updated estimates of target positions and the covariance matrix. Based on~\cite{boyd2004fastest}, it is guaranteed that once the consensus algorithm converges, each robot has the same estimates of targets' positions as its neighbors.

\section{Experiments}
Through simulated experiments in the Gazebo environment, we show that the proposed system can track multiple targets in a decentralized, risk-aware, and adaptive fashion, while achieving performance comparable to its centralized counterpart. 
    
    \subsection{Decentralized Tracking}
    \label{experiment_1}
    We demonstrate that our decentralized target tracking system is: 
    \begin{enumerate}[label=\arabic*., wide, leftmargin=*]
        \item Able to track multiple targets simultaneously, and has a reasonable division of labor among the robots; 
        \item Risk-aware and adaptive to the sensor configuration; 
        \item Able to remain connected.
    \end{enumerate}
    
    At the start of the target tracking process, each robot is equipped with three heterogeneous sensors which have the following measurement vectors: 
    \begin{equation}
        \label{sensor_model}
        \mathbf{h}_1 = \big[\begin{matrix}
          1 & 0
        \end{matrix}\big],
        \mathbf{h}_2 = \big[\begin{matrix}
          0 & 1
        \end{matrix}\big],
        \mathbf{h}_3 = \big[\begin{matrix}
          1 & 1
        \end{matrix}\big]
    \end{equation}
     \ie the first type of sensor has sensing ability in the $x$ direction, the second type of sensor has sensing ability in the $y$ direction, and the third type has both. Sensor failures are simulated randomly by flipping a biased coin, where the probability of failure is proportional to the total risk at each robot's position induced by all the targets. Each robot's random sensor failure is generated independently. For an illustration of the system performance, a demo of Gazebo simulations is available online\footnote{\url{https://www.youtube.com/watch?v=AAOKDGqkuz8}}, where five robots track three targets. 
     
     Here, we present qualitative results for the setting where five robots track four targets. The initial sensor status as described in Eq.~(\ref{eq:sensor_status}) is $\mathbf{\Gamma}_0 = \mathbf{1}_{5 \times 3}$. For each robot $i \in \{1, 2,\cdots, 5\}$, we let the parameters to trade off between tracking quality maximization and risk minimization in Eq.~(\ref{eq:cost}) be $q_1 = 3.0, q_2 = 20.0$. The maximum tracking error in Eq.~(\ref{eq:perf}) is chosen as $\boldsymbol{\rho}_{i1} = [0.001; 0.001; 0.001; 0.001]$, and the maximum risk in Eq.~(\ref{eq:risk-aware}) is $\rho_{i2}=0.15$. The connectivity threshold in Eq.~(\ref{eq:qp_connect}) is chosen to be $\epsilon = 0.25$. 

    \begin{figure}
    \captionsetup[subfigure]{font=scriptsize,labelfont=scriptsize}
    \centering
          \subfloat[$R_\text{comm}=11.0$, Risk-agnostic\label{fig:traj_2}]{ \includegraphics[width=0.3\textwidth]{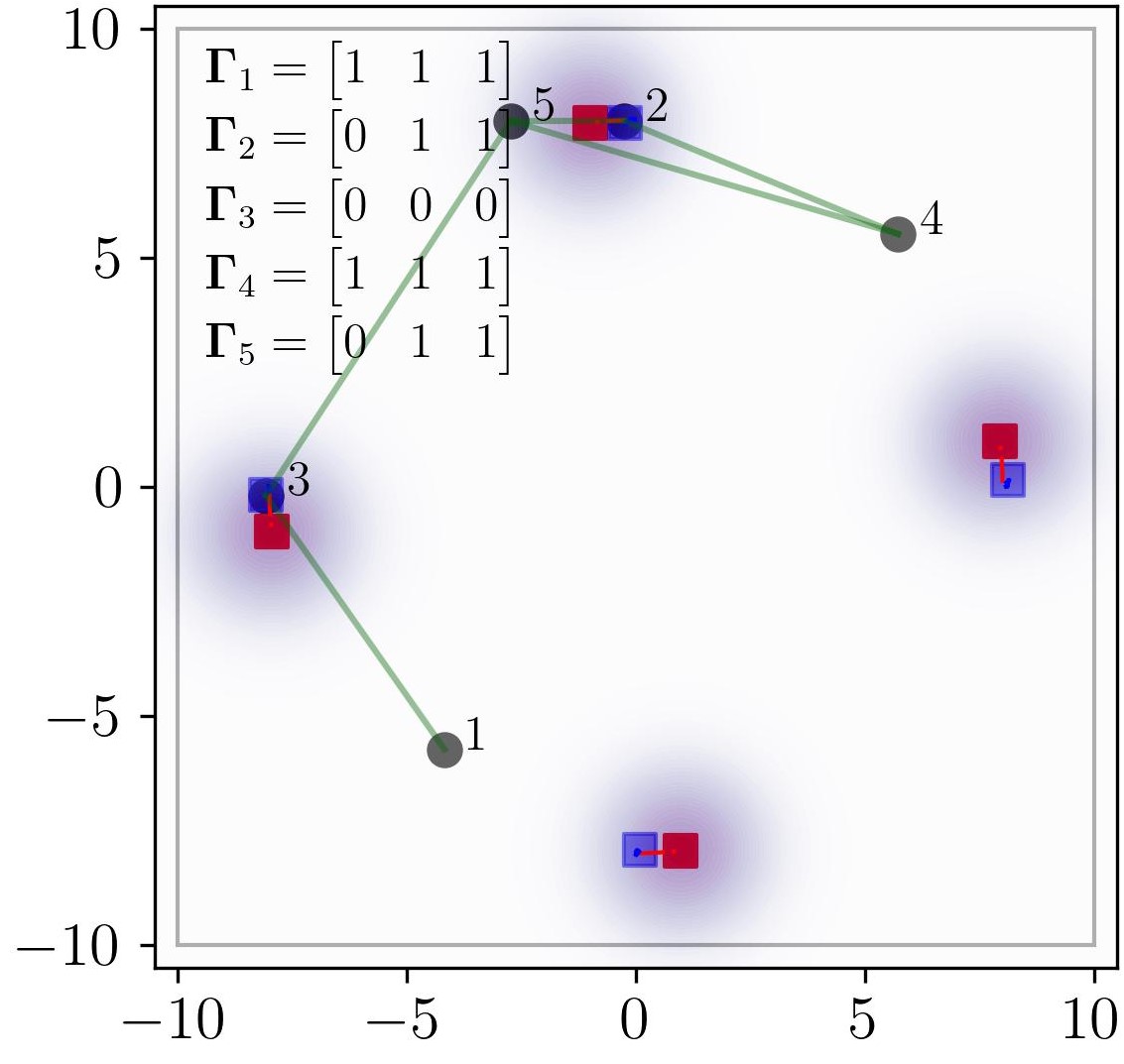}}
          \subfloat[$R_\text{comm}=11.0$, Risk-aware\label{fig:traj_1}]{ \includegraphics[width=0.3\textwidth]{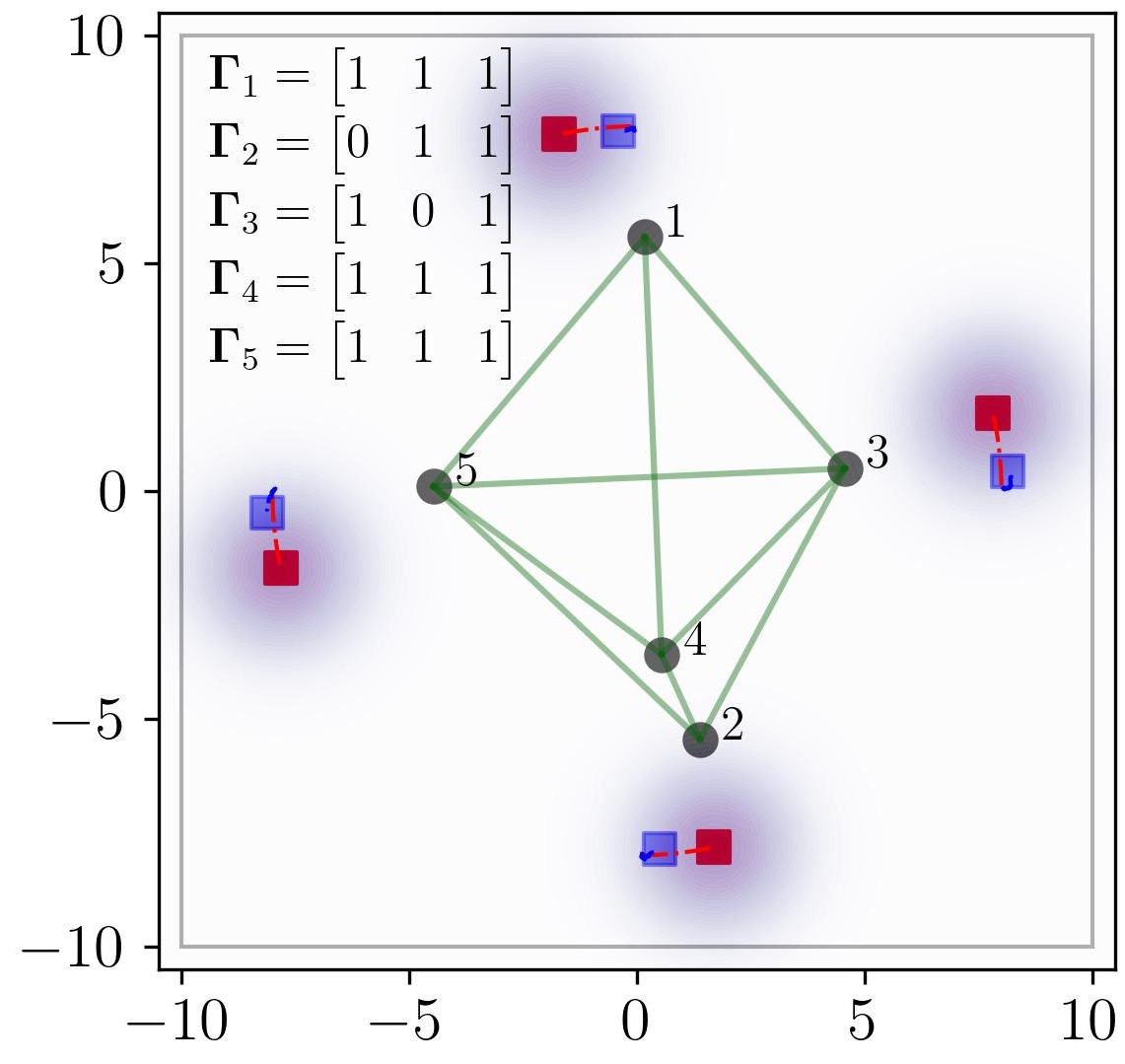}}
          \quad
          \subfloat[$R_\text{comm} = 7.0$, Risk-aware\label{fig:traj_3}]
          {\includegraphics[width=0.3\textwidth]{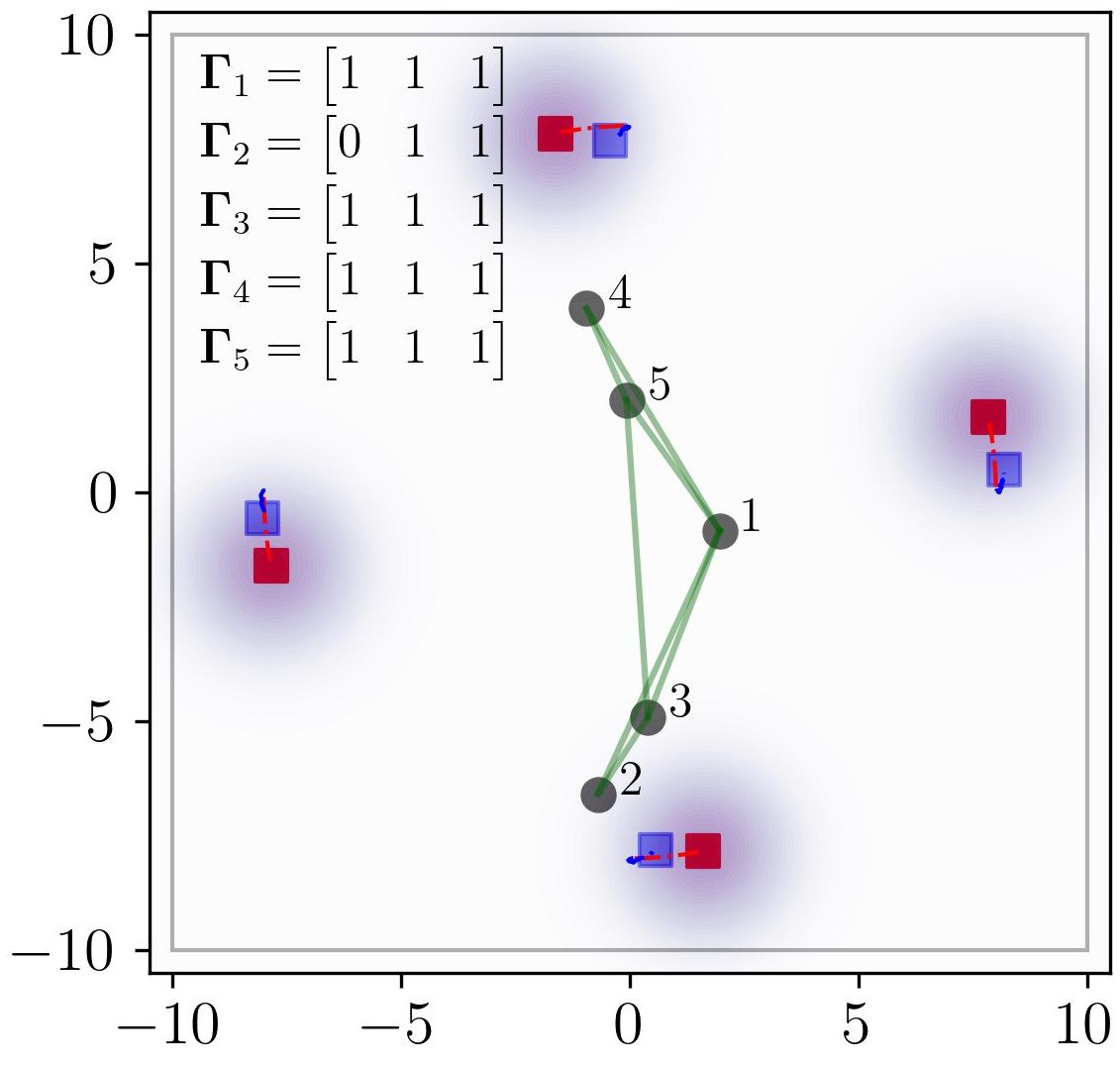}}
        \caption{Comparison of the target tracking behaviors under three different settings. Grey circles represent the robots. We draw true targets' positions and their trajectories in the latest 20 steps using red squares and red dotted lines. Correspondingly, blue squares and blue dotted lines represent the estimated targets' positions and trajectories. The purple-shaded regions around the targets indicate the distance-dependent risk field. The green edge denotes that the robots on its two endpoints are connected. }
        \label{fig:exp1-trajectory}
    \end{figure}
    \begin{figure}[ht]
    \captionsetup[subfigure]{font=scriptsize,labelfont=scriptsize}
        \centering
        \subfloat[Tracking Error\label{fig:de-track-erro}]{
        \includegraphics[width=0.45\textwidth]{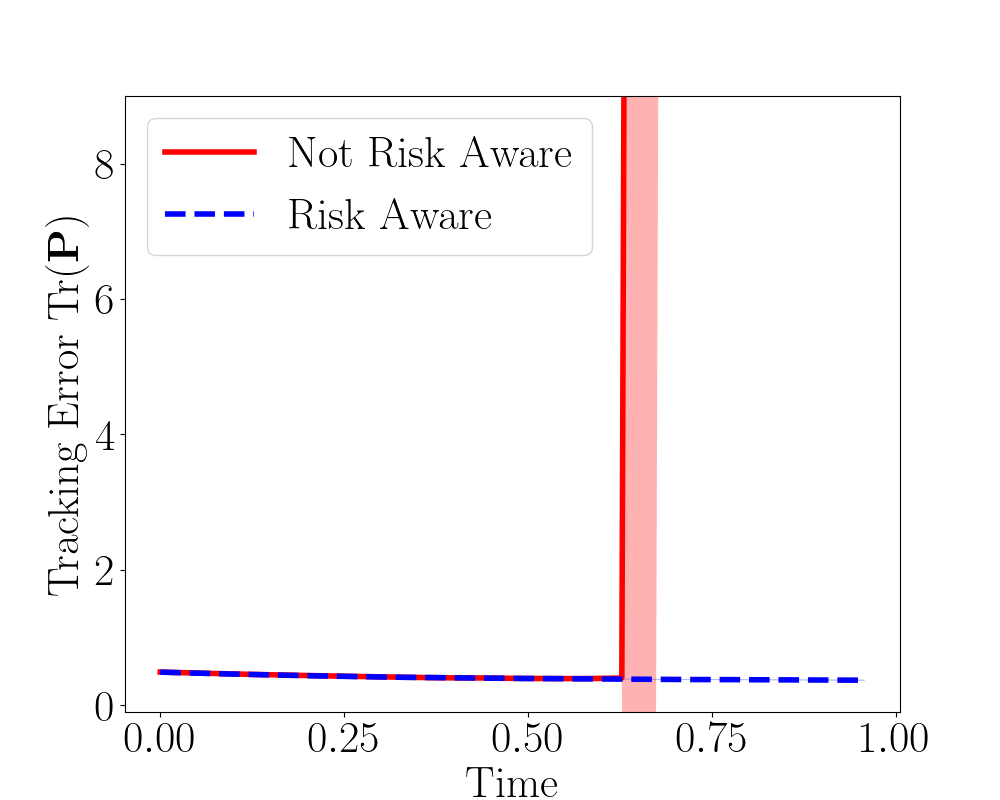}
        }\quad
        \subfloat[Sensor Margin\label{fig:de-sensor-margin}]{
          \includegraphics[width=0.45\textwidth]{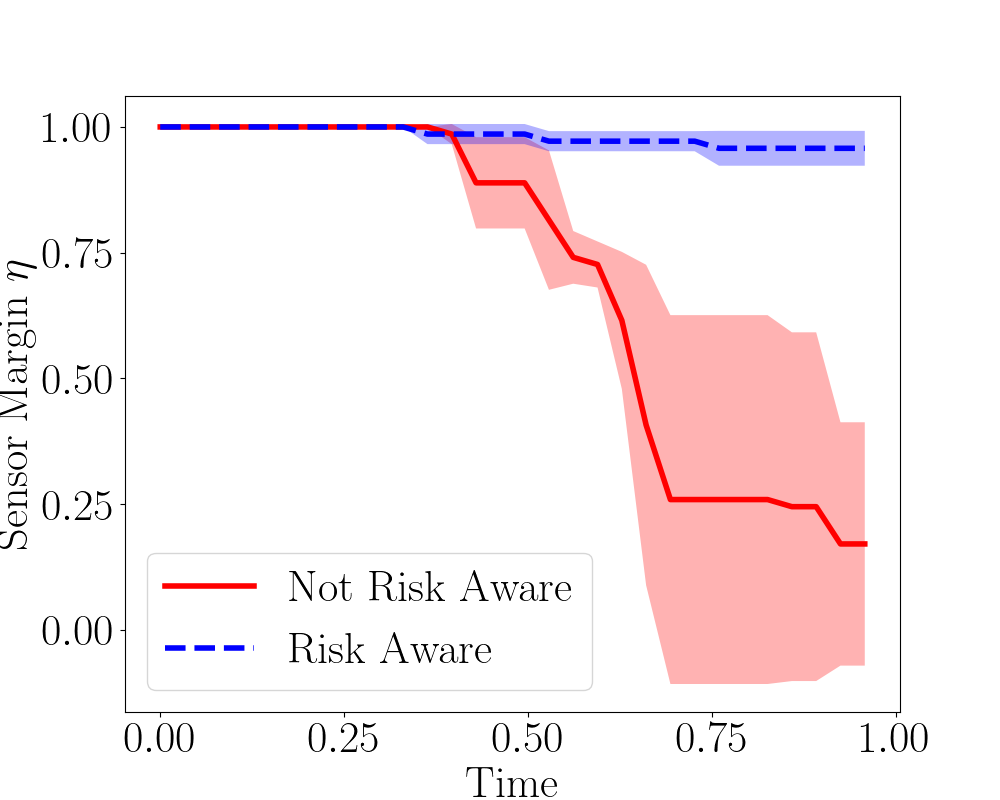}
        }
        \caption{Comparison of tracking error in Fig.~\ref{fig:de-track-erro} and sensor margin in Fig.~\ref{fig:de-sensor-margin} when the risk-aware constraint is vs is-NOT added to our decentralized target tracking framework with $R_\text{comm}=18.0$.}
        \label{fig:ens-compare}
    \end{figure}
    \begin{figure}[htp] 
        \centering
        \includegraphics[width=0.9\textwidth]{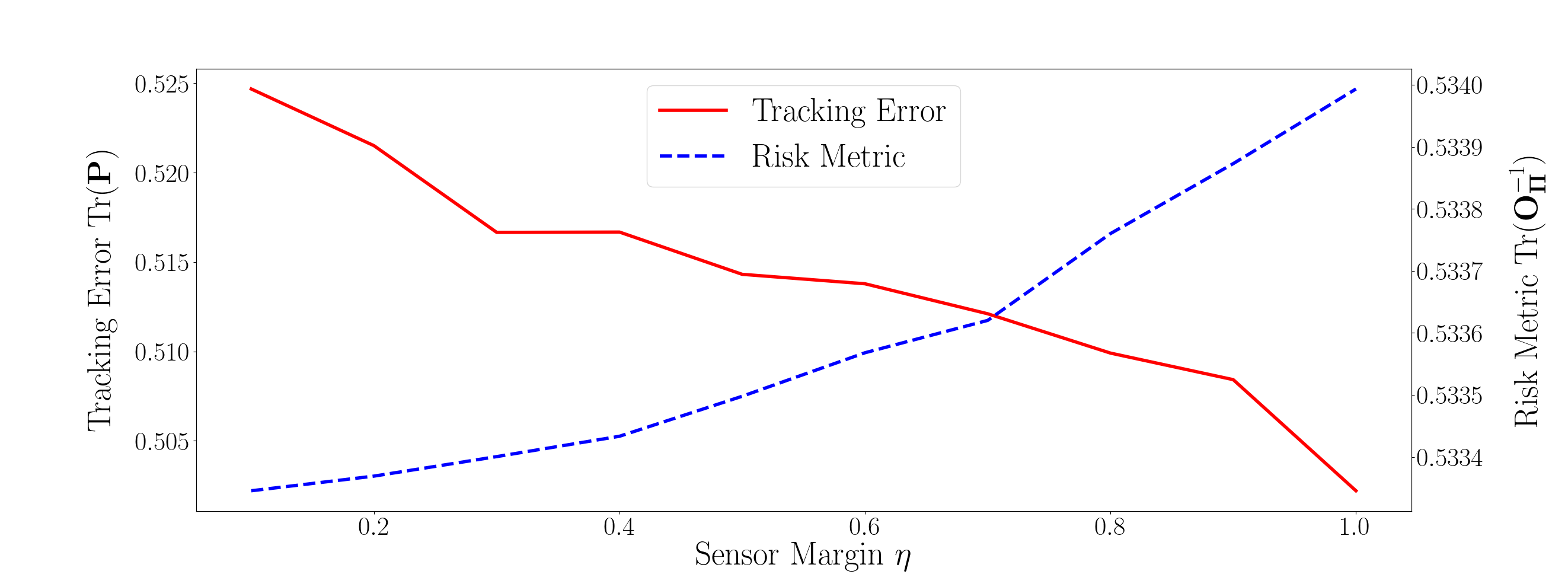}
        \caption{The trade-off between tracking error and risk level with respect to the sensor margin. When $\eta$ is small, the system prioritizes safety; as $\eta$ increases, the system prioritizes higher tracking accuracy. }
        \label{fig:double-compare}
    \end{figure}
    
    Fig.~\ref{fig:exp1-trajectory} compares the behavior of robots under three different settings: risk-agnostic with communication range $R_\text{comm} = 11.0$ (Fig.~\ref{fig:traj_2}), risk-aware with $R_\text{comm} = 11.0$ (Fig.~\ref{fig:traj_1}), and risk-aware with $R_\text{comm} = 7.0$ (Fig.~\ref{fig:traj_3}). To simulate the risk-agnostic case, we remove constraint Eq.~(\ref{eq:risk-aware}) and the second term in the cost function Eq.~(\ref{eq:cost}) for all robots. All robots are initially placed around the origin, and the targets move in concentric circles. In all three cases, the robots can move closer to the targets, so that they could obtain more accurate estimates of target positions. Meanwhile, these five robots spread out to track different targets. 
    
    Fig.~\ref{fig:traj_2} shows that without risk awareness, the robots directly approach the targets to get better estimates of the targets' positions. In contrast, the robots keep a certain distance from the targets if the risk-aware constraint is added, as shown in Fig.~\ref{fig:traj_1}. Such risk-aware behavior helps the robot team preserve its sensor suite for a longer time. Fig.~\ref{fig:traj_3} shows the behavior of the robot team when the communication range is relatively small. Compared to Fig.~\ref{fig:traj_1}, the robot team is more sparsely connected in this case.
    
    We further present quantitative results to show the effectiveness of risk awareness in the proposed decentralized system. We utilize two metrics, $\Tr(\mathbf{P})$ and $\eta$, which are the trace of the target state estimation covariance matrix and the sensor margin, respectively, both averaged across all robots. For both risk-aware and risk-agnostic cases, we run five trials and generate sensor failures randomly. Fig.~\ref{fig:ens-compare} shows the comparison for a 3-vs-3 case with $\boldsymbol{\rho}_{i1} = [0.01; 0.01; 0.01], \rho_{i2}=0.33, q_1 = 1.0, q_2 = 10.0, \forall i \in \{1, 2, 3\}$ in program~(\ref{non-convex-opt}), and $\epsilon = 0.25$ in Eq.~(\ref{eq:qp_connect}). The initial sensor configuration is $\mathbf{\Gamma}_{0} = \mathbf{1}_{3 \times 3}$. The communication range is $R_\text{comm} = 18.0$ so that the robots can easily maintain connectivity. It can be observed that if we remove the risk-aware constraint Eq.~(\ref{eq:risk-aware}), the sensor margin of the system will decrease dramatically because of more sensor failures. As a result, the tracking error upsurges sharply as the increasing number of failures quickly renders the system non-observable. 

    To demonstrate the system's ability to trade off between tracking quality and risk aversion, we compare $\text{Tr}(\mathbf{P})$ and $\text{Tr}(\mathbf{O}^{-1}_{\Pi})$ with respect to the sensor margin $\eta$. The result is presented in Fig.~\ref{fig:double-compare}. Note that, different from the above setting where the sensor margin is computed based on the available suite of sensors at each time step, here $\eta$ is specified as a sequence of fixed values to observe the system's behavior when the abundance of sensors is varying. As is expected, when the sensor margin $\eta$ is large, there are abundant sensors in the team and the system prioritizes tracking accuracy; conversely, as $\eta$ decreases, the system settles for a relatively higher tracking error to preserve its sensor suite. 
     
    \subsection{Comparison to Centralized Target Tracking}
    \begin{figure}[ht]
    \captionsetup[subfigure]{font=scriptsize,labelfont=scriptsize}
        \centering
          \subfloat[Tracking Error\label{fig:comp-track-error}]{\includegraphics[width=0.45\textwidth]{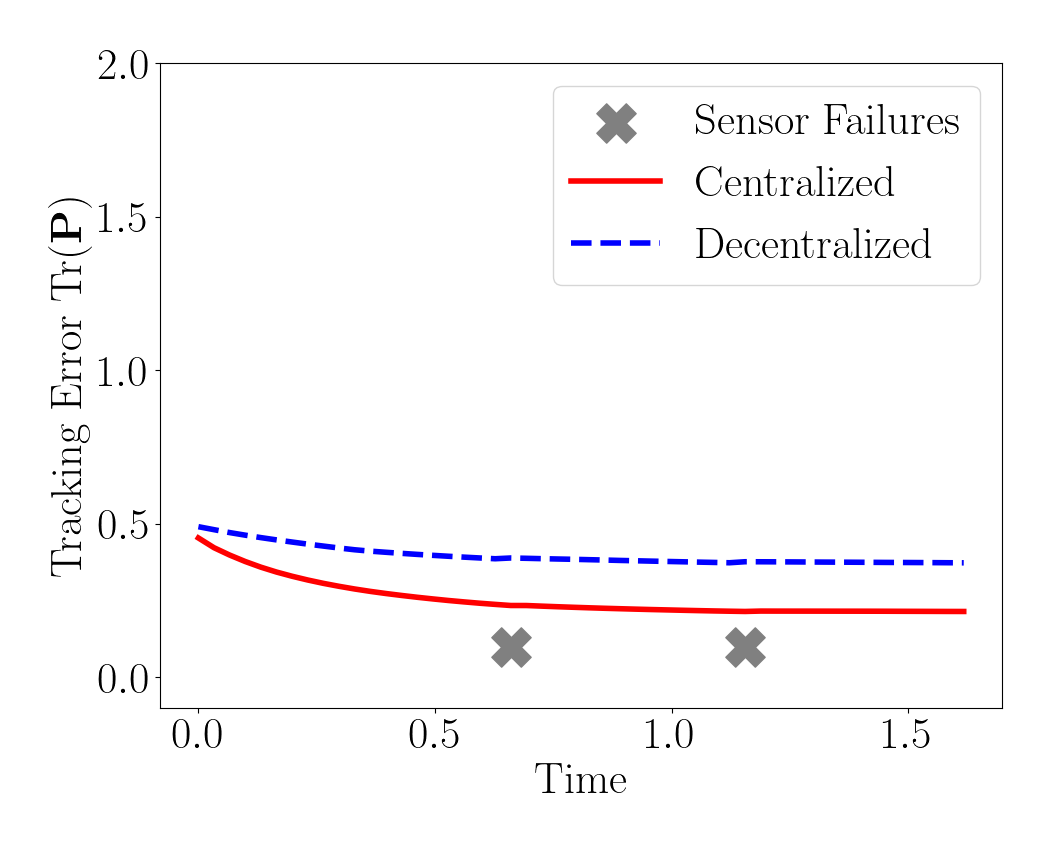}} \quad
          \subfloat[Sensor Margin\label{fig:comp-sensor-margin}]{\includegraphics[width=0.45\textwidth]{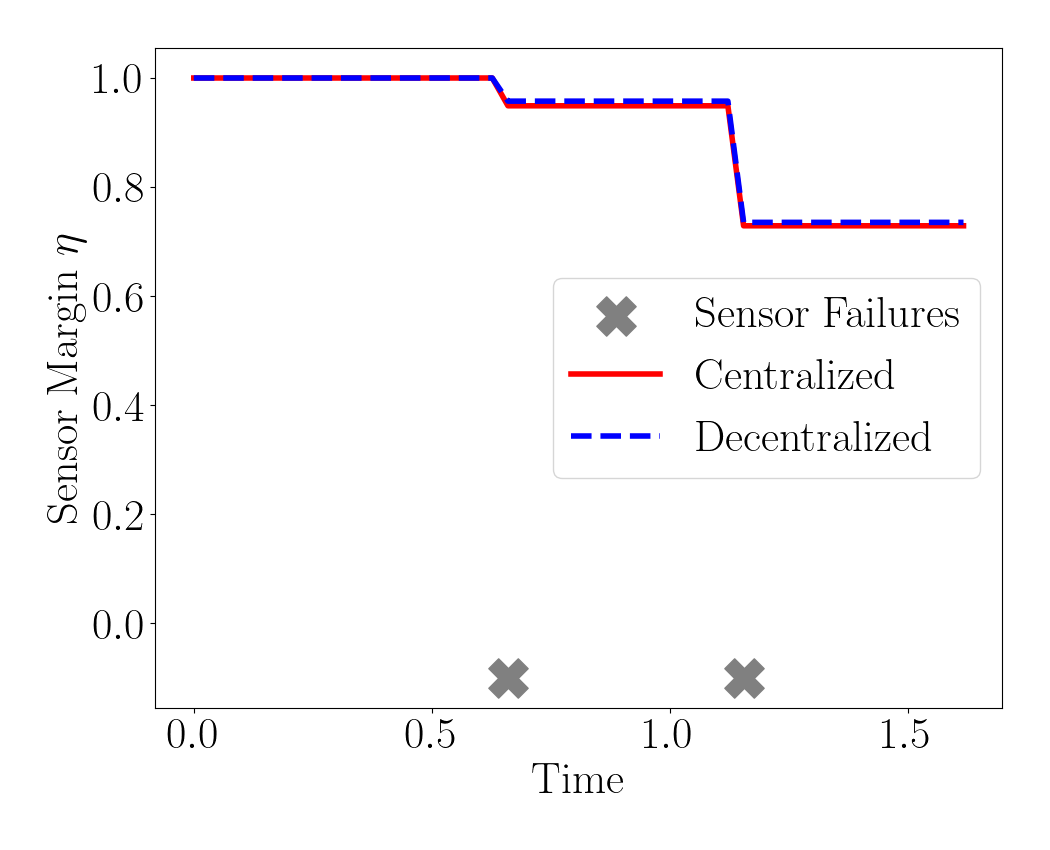}}
        \caption{Comparison of the centralized and decentralized systems under the same setting. We enforce the same sequence of sensor failures on both of them as denoted by the grey crosses. For the decentralized system, the $\Tr(\mathbf{P})$ and $\eta$ shown are both obtained from averaging across all robots. }
        \label{fig:de-cen-compare}
    \end{figure}
    We compare the proposed decentralized target tracking system with its centralized counterpart in terms of tracking error and sensor margin. The centralized tracking system used for comparison is from our previous work~\cite{mayya2022adaptive}. For a fair comparison, we use the same CBFs  (Eq.~(\ref{eq:qp_connect}) \&~(\ref{eq:qp_collision_avoid})) for network connectivity maintenance and collision avoidance. 
    
    We consider the scenario where 4 robots are tasked to track 4 targets. In both the centralized and decentralized systems, we let $R_\text{comm} = 18.0$, $q_1 = 1.0, q_2 = 10.0$ in Eq.~(\ref{eq:cost}), and the connectivity threshold in Eq.~(\ref{eq:qp_connect}) be $\epsilon = 0.25$. The maximum tracking error in Eq.~(\ref{eq:perf}) is chosen as $\boldsymbol{\rho}_{i1, j} = 0.01, \forall j\in \{1, 2, 3, 4\}$, and the maximum risk level in Eq.~(\ref{eq:risk-aware}) is chosen as $\rho_{i2}=0.33$, for each robot indexed $i\in \{1, 2, 3, 4\}$. Initially, each robot is equipped with three sensors with measurement matrices $\mathbf{h}_1, \mathbf{h}_2, \mathbf{h_3}$, respectively, \ie the initial sensor configuration as described in Eq.~(\ref{eq:sensor_status}) is $\mathbf{\Gamma}_0 = \mathbf{1}_{4\times 3}$. To conveniently compare their performance, we enforce the same sequence of pre-specified sensor failures instead of inducing sensor failures randomly. In a total of 50 time steps, a sensor whose measurement matrix is equal to $\mathbf{h}_3$ is damaged at time step 20; another sensor whose measurement matrix is $\mathbf{h}_2$ is damaged at step 35. The comparison of the tracking error and sensor margin is illustrated in Fig.~\ref{fig:de-cen-compare}. When both systems have the same changes in their sensor margin as shown in Fig.~\ref{fig:comp-sensor-margin}, the tracking error of the decentralized system is similar to that of the centralized one as shown in Fig.~\ref{fig:comp-track-error}.

\section{Conclusion and Future Work}
In this paper, we designed a decentralized and risk-aware multi-target tracking system. This is motivated by the setting where a team of robots is tasked with tracking multiple adversarial targets and approaching the targets for more accurate target position estimation increases the risk of sensor failures. We develop a decentralized approach to address the trade-off between tracking quality and safety. Each robot plans its motion to balance conflicting objectives -- tracking accuracy and safety, relying on its own information and communication with its neighbors. We utilize control barrier functions to guarantee network connectivity throughout the tracking process. We show via simulated experiments that our system achieves similar tracking accuracy and risk-awareness to its centralized counterpart. Since our focus is the framework design, we adopted a simplified linear measurement model. Therefore, one future research direction is to utilize more realistic observation models, \eg range and bearing sensors~\cite{zhou2011multirobot}, which requires linearized approximations in the optimization program. A second future direction relates to the computation of goal positions. In this work, the robots generate ideal goal positions by solving a non-convex optimization program. This gradient-based approach is in general computationally time-consuming. When a very strict requirement is specified for the tracking accuracy, it is possible that the system would fail to find the global optimal solution and get stuck in local optima. To this end, we plan to design more efficient distributed algorithms to compute the ideal goal positions~\cite{atanasov2015decentralized}. Additionally, we consider extending the proposed framework to deal with the scenarios where the targets can act strategically to compromise the sensors and communications of the robots~\cite{zhou2018resilient,shi2020robust,ramachandran2020resilient,liu2021distributed,zhou2022distributed,zhou2022robust}.

%
%
\bibliography{references}


\end{document}